\pdfoutput=1
\documentclass[11pt]{article}
\usepackage[a4paper,margin=26mm,headheight=14pt]{geometry}
\ifdefined\XeTeXversion
  \usepackage{fontspec}
  \IfFontExistsTF{Times New Roman}{\setmainfont{Times New Roman}}{\setmainfont{TeX Gyre Termes}}
\else
  \usepackage[utf8]{inputenc}
  \usepackage[T1]{fontenc}
  \IfFileExists{newtxtext.sty}{\usepackage{newtxtext}}{\usepackage{mathptmx}}
\fi
\usepackage{amsmath}
\usepackage{graphicx}
\usepackage{booktabs}
\usepackage{xcolor}
\usepackage{tikz}
\usetikzlibrary{arrows.meta,positioning}
\usepackage{fancyhdr}
\usepackage[numbers,sort&compress]{natbib}
\usepackage[colorlinks=true,linkcolor=black,citecolor=black,urlcolor=black,hyperfootnotes=false]{hyperref}

\newcommand{\DatasetName}{ZEGO}
\newcommand{\PaperTitleEN}{\DatasetName: Large-Scale Egocentric Human Manipulation\protect\\ Video Curation from Unconstrained Web Videos}
\newcommand{\PaperTitleENPlain}{ZEGO: Large-Scale Egocentric Human Manipulation Video Curation from Unconstrained Web Videos}
\newcommand{\PaperAuthors}{Qiushen Pan\quad Haoming Zhao\quad Chen Lin\quad Song Ran\quad Ruicheng Zhong\textsuperscript{*}}
\newcommand{\PaperAuthorsPlain}{Qiushen Pan; Haoming Zhao; Chen Lin; Song Ran; Ruicheng Zhong}
\newcommand{\PaperAffiliation}{Zetta Tech Pte. Ltd.}
\newcommand{\PaperContactEmail}{contact@zatt.ai}

\fancypagestyle{plain}{
  \fancyhf{}
  \fancyfoot[C]{\small\thepage}
  
}
\hypersetup{pdfauthor={\PaperAuthorsPlain},pdftitle={\PaperTitleENPlain}}
\author{\PaperAuthors\\[0.35em]
  {\normalsize\PaperAffiliation}\\
  {\normalsize\href{mailto:\PaperContactEmail}{\PaperContactEmail}}}
\newcommand{\correspondingnote}[1]{%
  \begingroup
  \renewcommand{\thefootnote}{}%
  \footnotetext{#1}%
  \endgroup
}
\date{}
\title{\PaperTitleEN}
\hypersetup{pdftitle={\PaperTitleENPlain}}
\begin{document}
\maketitle
\correspondingnote{\textsuperscript{*}Corresponding author: Ruicheng Zhong (\href{mailto:\PaperContactEmail}{\PaperContactEmail}).}
\begin{abstract}
Large-scale egocentric human manipulation video provides essential physical interaction priors for embodied intelligence and vision-language-action (VLA) pretraining. Although unconstrained web videos span a vast spectrum of everyday chores, crafts, and technical repairs, genuinely clean continuous manipulation is exceedingly scarce---heavily diluted by non-manipulative recordings, camera-facing narration, and rapid editing cuts, while tool occlusions and motion blur cause single hand-perception models to suffer intermittent dropouts. We present a scalable multi-stage curation and automatic annotation methodology for mining egocentric manipulation clips from open web videos, together with the \DatasetName{} dataset. Across $\sim 230\text{M}$ candidate videos surfaced by $\sim 900\text{K}$ multilingual task queries, a sparse preview pre-screening gate evaluates unordered keyframe samples via heterogeneous hand-and-face voting prior to full-stream download to discard $\sim 96\%$ of deduplicated candidates upfront (admitting $\sim 4\%$ into full-video processing), followed by decode-level media integrity verification and neural shot-boundary detection ($\le 10\text{ s}$ cuts removed); next, over $1\text{ Hz}$ streams, it jointly runs cascaded palm-to-landmark regression and single-stage full-frame hand detection, performing same-frame instance association and confidence-weighted 2D keypoint fusion to repair occlusion-induced fragmentation alongside face-hand detachment, dual-track occlusion tolerance, and palm-scale-normalized kinematic trimming; finally, multimodal large models inspect temporally ordered frame sequences to reject virtual/CG scenes, vehicle driving, and residual non-first-person angles while generating structured English action, object, tool, and visual-overlay annotations. Controlled ablations show that heterogeneous hand fusion expands continuous yield by 30.7\% to 238.5\% over individual models. Across 6.66 million admitted deduplicated videos (2.07 million hours), full-stream front-end curation prunes a further 84.84\% of non-target duration and extracts 314.5 thousand hours of dynamic candidates at 89.71\% unified first-person precision ($\approx 0.54\%$ compound end-to-end yield). \DatasetName{} comprises 5,117,314 high-definition ($\ge 1080\text{p}$) manipulation clips totaling 39,400.16 hours from 686,914 source videos across over 400,000 action and object expressions, including \DatasetName{}-Clean, a 2,583,138-clip (21,018.85-hour) watermark- and subtitle-free single-view subset.
\end{abstract}

\section{Introduction}
Embodied intelligence seeks to equip robots with generalizable manipulation skills in unstructured physical environments. Acquiring such dexterity demands broad, diverse hand-object interaction experience~\cite{openx2023,kim2024openvla}. Because on-robot teleoperation is bottlenecked by hardware availability and human operator labor, leveraging human egocentric manipulation videos without explicit robot action labels has emerged as a cornerstone for visual representation pretraining~\cite{nair2022r3m,xiao2022mvp,ma2023vip}, latent action modeling~\cite{ye2025lapa}, and world-action pretraining. Recent physical scaling studies further show that expanding human-video pretraining from $10^3$ to $10^6$ hours yields monotonic gains across diverse real-robot manipulation tasks under fixed downstream post-training~\cite{dyna2026dyna2}. Motivated by this data-infrastructure bottleneck, this paper studies an algorithmic and curation problem: how to extract and annotate continuous, high-purity egocentric human manipulation episodes at scale from unconstrained open-web videos.

Organized wearable-camera capture---spanning household kitchens, multi-city participant cohorts, and laboratory head-mounted displays---remains the classical paradigm for building egocentric benchmarks~\cite{damen2022epickitchens,grauman2021ego4d,hoque2025egodex}. However, device logistics, battery life, and fixed recording sites typically cap curated manipulation volume at hundreds to a few thousand hours, limiting long-tail tool and environmental coverage~\cite{yang2026aoe}. Conversely, publicly accessible Internet videos already contain vast repositories of craft tutorials, mechanical and electronic repairs, cooking, and household organization, many intervals of which naturally exhibit an egocentric viewpoint, clear hand-object contact, and uninterrupted motion. Yet raw web videos are far from training-ready: (1) a single recording routinely alternates between hands-on manipulation, face-to-camera narration, split-screen composites, and transition B-roll; (2) visible hands frequently reflect conversational gesturing, passive stationary object display, or rendered video-game/CG avatars rather than physical manipulation; and (3) tool occlusions, rapid motion blur, and image-border truncation cause individual hand detectors to drop detections for several seconds, fragmenting or discarding valid continuous episodes.

To address these challenges, we propose a multi-stage curation and annotation framework that couples sparse preview pre-screening, full-stream physical-geometric concentration, and multimodal semantic verification, and construct \DatasetName{}, a large-scale open egocentric human manipulation dataset. Across $\sim 230\text{M}$ candidate videos queried via $\sim 900\text{K}$ multilingual search keywords, an entry pre-screening gate evaluates sparse preview keyframes via heterogeneous hand-and-face voting prior to full-video download to discard $\sim 96\%$ of deduplicated web candidates upfront. Across the admitted $\sim 4\%$ full-length videos, the front end applies decode-level integrity verification, neural shot-boundary partitioning, and short-shot pruning ($>10\text{ s}$), followed by parallel $1\text{ Hz}$ inference across cascaded palm-to-landmark regression and single-stage multi-scale full-frame hand detection. Through cross-model merged-hand suppression, same-frame instance association, confidence-weighted 2D keypoint fusion, face-hand geometric detachment, dual-track occlusion tolerance, and palm-scale-normalized kinematic trimming, the front end isolates compact dynamic manipulation intervals. At the back end, vision-language models inspect temporally ordered frame sequences to filter out virtual/CG scenes, vehicle driving, and residual non-first-person angles while producing open-vocabulary English annotations and visual-overlay flags. Our contributions are threefold:
\begin{itemize}
    \item \textbf{Multi-Stage Cascade Curation with Complementary Hand Fusion:} We combine sparse preview pre-screening with instance-level cross-model association, confidence-weighted 2D keypoint fusion, and palm-scale-normalized dual-scale kinematic trimming. Controlled ablations show 30.7\%--238.5\% higher continuous yield over single detectors, while large-scale deployment prunes 84.84\% of non-target duration across 2.07 million hours of preview-admitted videos at 89.71\% candidate precision---quantifying and overcoming the $\approx 0.54\%$ ($\approx 0.44\%$ overlay-free) end-to-end scarcity of clean egocentric manipulation on the open web.
    \item \textbf{The \DatasetName{} Multi-Tier Dataset:} We release 5,117,314 high-definition ($\ge 1080\text{p}$) continuous manipulation clips totaling 39,400.16 hours from 686,914 unique source videos---exceeding existing dedicated egocentric manipulation benchmarks by more than an order of magnitude---alongside explicit watermark/subtitle metadata and \DatasetName{}-Clean, a 21,018.85-hour (2,583,138-clip) overlay-free single-view subset tailored for visual world models.
    \item \textbf{Open-Vocabulary Semantic and 2D Trajectory Annotations:} Every clip provides $1\text{ Hz}$ fused 21-joint 2D hand observations together with sequential English captions, primary actions (401.6K phrases), manipulated objects (404.4K expressions), and external tools (21.1K types), backed by decoupled observation-lineage storage for reproducible policy evolution.
\end{itemize}

\section{Related Work}
\paragraph{Human-Video Pretraining for Embodied Intelligence.}
Joint training across multi-robot demonstration repositories shows that scaling environmental and object diversity improves policy generalization~\cite{openx2023,kim2024openvla}. Vision-language-action (VLA) architectures such as RT-2~\cite{brohan2023rt2} and $\pi_0$~\cite{black2024pi0} further transfer web-scale semantic priors to physical robot control. To overcome the scarcity of teleoperated trajectories, R3M~\cite{nair2022r3m}, MVP~\cite{xiao2022mvp}, and VIP~\cite{ma2023vip} pretrain generalizable visual encoders on human egocentric manipulation videos, while LAPA~\cite{ye2025lapa} extracts unsupervised latent action tokens from action-free human videos. Recently, the Dyna-2 technical report~\cite{dyna2026dyna2} evaluated human-video pretraining scales from $10^3$ to $10^6$ hours with downstream robot post-training held fixed, improving mean normalized performance across 14 physical manipulation tasks from 20\% to 53\% and highlighting the need for tens to hundreds of thousands of hours of clean egocentric manipulation footage.

\paragraph{Egocentric and Hand-Object Interaction Datasets.}
Dedicated wearable capture is the primary avenue for constructing egocentric benchmarks. EPIC-KITCHENS-100~\cite{damen2022epickitchens} records 100 hours of unscripted kitchen activities; HOI4D~\cite{liu2022hoi4d} and Assembly101~\cite{sener2022assembly101} capture indoor tabletop object interactions and toy assembly; Ego4D~\cite{grauman2021ego4d} and Ego-Exo4D~\cite{grauman2024egoexo4d} expand collection across multi-city daily and skilled domains (3,670 hours and 221 hours of egocentric video, respectively) while documenting demographic and battery-life sampling constraints; EgoDex~\cite{hoque2025egodex} collects 829 hours of tabletop dexterous manipulation with head-mounted 3D tracking; and Something-Something V2~\cite{goyal2017something} crowdsources short common-sense object-interaction clips. While dedicated capture offers controlled sensor synchronization, hardware deployment and prescribed tabletop setups constrain continuous scaling across open-world tools and environments~\cite{yang2026aoe}.

\paragraph{Web Video Curation and Hand Perception.}
Open Internet repositories offer immense task breadth: HowTo100M~\cite{miech2019howto100m} compiles narrated instructional videos across 23,000 tasks, 100DOH~\cite{shan2020100doh} annotates frame-level hand-object contact boxes on web footage, and Panda-70M~\cite{chen2024panda70m} and InternVid~\cite{wang2024internvid} curate general video-text pairs via shot splitting and multimodal captioning. However, general video-text corpora do not enforce egocentric viewpoints or hand-manipulation continuity, whereas static frame benchmarks lack uninterrupted temporal trajectories. In hand perception, cascaded palm-detection-and-cropped-regression models such as MediaPipe Hands~\cite{zhang2020mediapipehands} deliver precise joint localization in clear views, yet lose entire hand instances when tools occlude the palm, rapid motion induces blur, or hands enter partially at image borders. Conversely, single-stage multi-scale full-frame hand detection and 2D keypoint estimation branches---such as the localization backbone of WiLoR~\cite{potamias2025wilor}---achieve substantially higher recall across scale shifts and clutter without incurring the heavy cost of 3D mesh recovery~\cite{romero2017mano} during candidate screening. Our framework unifies heterogeneous 2D hand perception, physical kinematic trimming, and multimodal semantic verification for scalable egocentric manipulation curation.

\section{Multi-Stage Curation and Annotation Methodology}
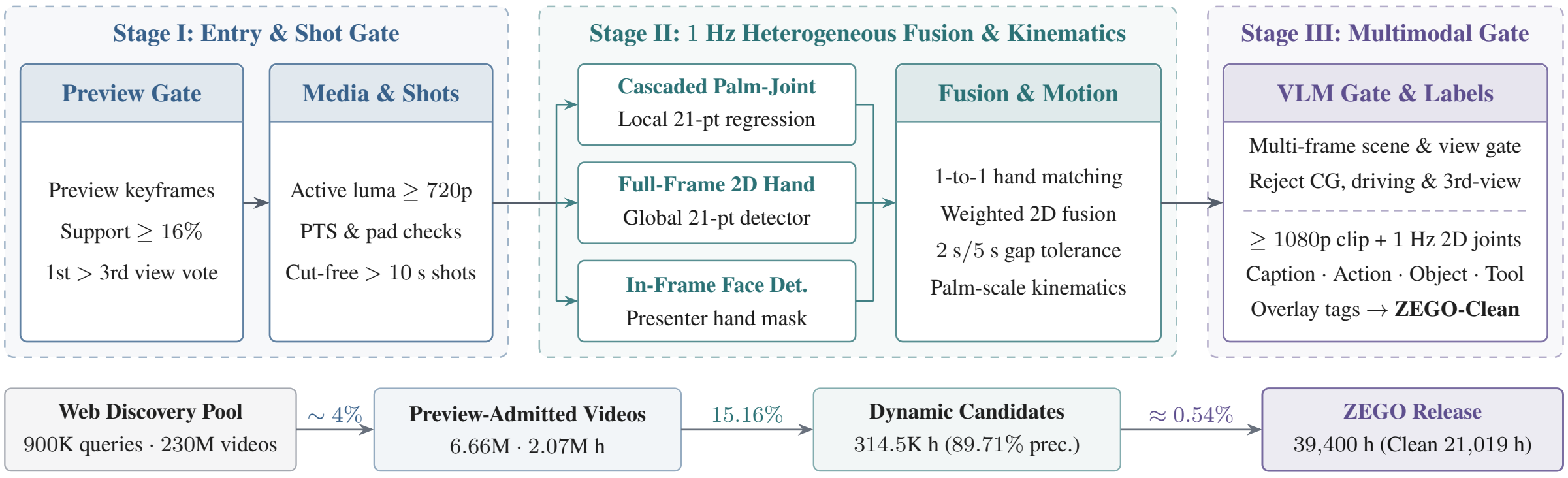
\begin{figure}[htbp]
\centering
\definecolor{GateBlue}{RGB}{44,88,132}
\definecolor{PercTeal}{RGB}{32,106,110}
\definecolor{SemIndigo}{RGB}{86,70,136}
\definecolor{FunnelSlate}{RGB}{55,65,81}
\resizebox{\textwidth}{!}{%
\begin{tikzpicture}[
    x=1mm, y=1mm,
    arr/.style={-{Stealth[length=2.1mm,width=1.55mm]}, line width=0.7pt, draw=FunnelSlate!85},
    subarr/.style={-{Stealth[length=1.8mm,width=1.35mm]}, line width=0.6pt, draw=PercTeal!85},
    farr/.style={-{Stealth[length=2.0mm,width=1.5mm]}, line width=0.65pt, draw=FunnelSlate!75}
]
\filldraw[fill=GateBlue!4, draw=GateBlue!45, dashed, rounded corners=3.4pt, line width=0.6pt]
    (0.0,14.6) rectangle (64.6,59.6);
\node[font=\footnotesize\bfseries, text=GateBlue!95] at (32.3,55.9) {Stage I: Entry \& Shot Gate};

\filldraw[fill=white, draw=GateBlue!75, rounded corners=2.5pt, line width=0.65pt]
    (2.0,16.6) rectangle (30.6,52.2);
\fill[GateBlue!14, rounded corners=2.5pt] (2.15,44.8) rectangle (30.45,52.05);
\fill[GateBlue!14] (2.15,44.8) rectangle (30.45,47.0);
\draw[GateBlue!65, line width=0.55pt] (2.0,44.8) -- (30.6,44.8);
\node[font=\footnotesize\bfseries, text=GateBlue!95] at (16.3,48.5) {Preview Gate};
\node[font=\scriptsize, align=center, text=black!88] at (16.3,30.7) {
    Preview keyframes\\[4.6pt]
    Support $\ge 16\%$\\[4.6pt]
    1st $>$ 3rd view vote
};

\draw[arr] (30.6,34.4) -- (34.0,34.4);

\filldraw[fill=white, draw=GateBlue!75, rounded corners=2.5pt, line width=0.65pt]
    (34.0,16.6) rectangle (62.6,52.2);
\fill[GateBlue!14, rounded corners=2.5pt] (34.15,44.8) rectangle (62.45,52.05);
\fill[GateBlue!14] (34.15,44.8) rectangle (62.45,47.0);
\draw[GateBlue!65, line width=0.55pt] (34.0,44.8) -- (62.6,44.8);
\node[font=\footnotesize\bfseries, text=GateBlue!95] at (48.3,48.5) {Media \& Shots};
\node[font=\scriptsize, align=center, text=black!88] at (48.3,30.7) {
    Active luma $\ge 720\text{p}$\\[4.6pt]
    PTS \& pad checks\\[4.6pt]
    Cut-free $>10\text{ s}$ shots
};

\filldraw[fill=PercTeal!4, draw=PercTeal!45, dashed, rounded corners=3.4pt, line width=0.6pt]
    (68.6,14.6) rectangle (150.4,59.6);
\node[font=\footnotesize\bfseries, text=PercTeal!95] at (109.5,55.9) {Stage II: $1\text{ Hz}$ Heterogeneous Fusion \& Kinematics};

\draw[line width=0.7pt, draw=FunnelSlate!85] (62.6,34.4) -- (70.8,34.4);
\draw[line width=0.6pt, draw=PercTeal!85] (70.8,21.8) -- (70.8,47.0);
\draw[subarr] (70.8,47.0) -- (73.6,47.0);
\draw[subarr] (70.8,34.4) -- (73.6,34.4);
\draw[subarr] (70.8,21.8) -- (73.6,21.8);

\filldraw[fill=white, draw=PercTeal!75, rounded corners=2.2pt, line width=0.6pt]
    (73.6,41.8) rectangle (109.2,52.2);
\node[font=\scriptsize, align=center, text=black!90] at (91.4,47.0) {
    \textbf{\color{PercTeal!95}Cascaded Palm-Joint}\\[1.6pt]
    Local 21-pt regression
};

\filldraw[fill=white, draw=PercTeal!75, rounded corners=2.2pt, line width=0.6pt]
    (73.6,29.2) rectangle (109.2,39.6);
\node[font=\scriptsize, align=center, text=black!90] at (91.4,34.4) {
    \textbf{\color{PercTeal!95}Full-Frame 2D Hand}\\[1.6pt]
    Global 21-pt detector
};

\filldraw[fill=white, draw=PercTeal!75, rounded corners=2.2pt, line width=0.6pt]
    (73.6,16.6) rectangle (109.2,27.0);
\node[font=\scriptsize, align=center, text=black!90] at (91.4,21.8) {
    \textbf{\color{PercTeal!95}In-Frame Face Det.}\\[1.6pt]
    Presenter hand mask
};

\draw[line width=0.6pt, draw=PercTeal!85] (109.2,47.0) -- (111.5,47.0);
\draw[line width=0.6pt, draw=PercTeal!85] (109.2,34.4) -- (111.5,34.4);
\draw[line width=0.6pt, draw=PercTeal!85] (109.2,21.8) -- (111.5,21.8);
\draw[line width=0.6pt, draw=PercTeal!85] (111.5,21.8) -- (111.5,47.0);
\draw[subarr] (111.5,34.4) -- (114.4,34.4);

\filldraw[fill=white, draw=PercTeal!75, rounded corners=2.5pt, line width=0.65pt]
    (114.4,16.6) rectangle (148.4,52.2);
\fill[PercTeal!14, rounded corners=2.5pt] (114.55,44.8) rectangle (148.25,52.05);
\fill[PercTeal!14] (114.55,44.8) rectangle (148.25,47.0);
\draw[PercTeal!65, line width=0.55pt] (114.4,44.8) -- (148.4,44.8);
\node[font=\footnotesize\bfseries, text=PercTeal!95] at (131.4,48.5) {Fusion \& Motion};
\node[font=\scriptsize, align=center, text=black!88] at (131.4,30.7) {
    1-to-1 hand matching\\[3.0pt]
    Weighted 2D fusion\\[3.0pt]
    $2\text{ s}/5\text{ s}$ gap tolerance\\[3.0pt]
    Palm-scale kinematics
};

\filldraw[fill=SemIndigo!4, draw=SemIndigo!45, dashed, rounded corners=3.4pt, line width=0.6pt]
    (154.4,14.6) rectangle (200.0,59.6);
\node[font=\footnotesize\bfseries, text=SemIndigo!95] at (177.2,55.9) {Stage III: Multimodal Gate};

\draw[arr] (148.4,34.4) -- (156.4,34.4);

\filldraw[fill=white, draw=SemIndigo!75, rounded corners=2.5pt, line width=0.65pt]
    (156.4,16.6) rectangle (198.0,52.2);
\fill[SemIndigo!14, rounded corners=2.5pt] (156.55,44.8) rectangle (197.85,52.05);
\fill[SemIndigo!14] (156.55,44.8) rectangle (197.85,47.0);
\draw[SemIndigo!65, line width=0.55pt] (156.4,44.8) -- (198.0,44.8);
\node[font=\footnotesize\bfseries, text=SemIndigo!95] at (177.2,48.5) {VLM Gate \& Labels};
\node[font=\scriptsize, align=center, text=black!88] at (177.2,39.2) {
    Multi-frame scene \& view gate\\[2.4pt]
    Reject CG, driving \& 3rd-view
};
\draw[SemIndigo!45, densely dashed, line width=0.55pt] (159.0,33.4) -- (195.4,33.4);
\node[font=\scriptsize, align=center, text=black!88] at (177.2,25.0) {
    $\ge 1080\text{p}$ clip + $1\text{ Hz}$ 2D joints\\[2.2pt]
    Caption $\cdot$ Action $\cdot$ Object $\cdot$ Tool\\[2.2pt]
    Overlay tags $\rightarrow$ \textbf{\DatasetName{}-Clean}
};

\filldraw[fill=FunnelSlate!5, draw=FunnelSlate!45, rounded corners=2.4pt, line width=0.55pt]
    (0.0,0.0) rectangle (37.4,10.6);
\node[font=\scriptsize, align=center, text=black!88] at (18.7,5.3) {
    \textbf{Web Discovery Pool}\\[1.4pt]
    $900\text{K}$ queries $\cdot$ $230\text{M}$ videos
};

\draw[farr] (37.4,5.3) -- node[above, font=\scriptsize\bfseries, text=GateBlue!95, inner sep=2pt] {$\sim 4\%$} (47.4,5.3);

\filldraw[fill=GateBlue!6, draw=GateBlue!50, rounded corners=2.4pt, line width=0.55pt]
    (47.4,0.0) rectangle (86.8,10.6);
\node[font=\scriptsize, align=center, text=black!88] at (67.1,5.3) {
    \textbf{Preview-Admitted Videos}\\[1.4pt]
    $6.66\text{M}$ $\cdot$ $2.07\text{M}$ h
};

\draw[farr] (86.8,5.3) -- node[above, font=\scriptsize\bfseries, text=PercTeal!95, inner sep=2pt] {$15.16\%$} (103.8,5.3);

\filldraw[fill=PercTeal!6, draw=PercTeal!55, rounded corners=2.4pt, line width=0.55pt]
    (103.8,0.0) rectangle (143.2,10.6);
\node[font=\scriptsize, align=center, text=black!88] at (123.5,5.3) {
    \textbf{Dynamic Candidates}\\[1.4pt]
    $314.5\text{K}$ h ($89.71\%$ prec.)
};

\draw[farr] (143.2,5.3) -- node[above, font=\scriptsize\bfseries, text=SemIndigo!95, inner sep=2pt] {$\approx 0.54\%$} (161.4,5.3);

\filldraw[fill=SemIndigo!11, draw=SemIndigo!80, rounded corners=2.4pt, line width=0.7pt]
    (161.4,0.0) rectangle (200.0,10.6);
\node[font=\scriptsize, align=center, text=black!92] at (180.7,5.3) {
    \textbf{\color{SemIndigo!95}\DatasetName{} Release}\\[1.4pt]
    $39{,}400\text{ h}$ (Clean $21{,}019\text{ h}$)
};
\end{tikzpicture}%
}
\caption{Multi-stage curation architecture and empirical scale-concentration funnel. Top: coordinated three-stage pipeline coupling sparse preview pre-screening, $1\text{ Hz}$ three-branch heterogeneous perception fusion and kinematic trimming, and multimodal semantic verification. Bottom: end-to-end distillation trajectory from $\sim 230\text{M}$ queried videos ($\sim 900\text{K}$ search keywords) to $\approx 0.54\%$ net retention.}
\label{fig:pipeline}
\end{figure}

Our curation methodology narrows unconstrained web videos progressively through a three-stage physical-to-semantic cascade (Figure~\ref{fig:pipeline}): upstream sparse preview pre-screening and media-shot gating, $1\text{ Hz}$ heterogeneous multi-branch perception fusion with physical kinematic trimming, and multimodal large-model semantic verification and structured annotation.

\subsection{Problem Formulation, Sparse Preview Pre-Screening, and Viewpoint Criterion}
We systematically expand approximately 900,000 multilingual search keywords across manipulation-dense domains---including handcrafts, electronics and mechanical repair, cooking, gardening, cleaning, and tabletop assembly---querying roughly 230 million candidate videos on public platforms. Even under fine-grained task-targeted queries, raw search results exhibit cross-query overlap and are overwhelmingly dominated by talking-head vlogs, slideshows, distant spectator footage, and static product displays. To avoid full-resolution stream acquisition across hundreds of millions of non-manipulative URLs, after platform-ID deduplication our pipeline enforces a \textbf{video-level sparse preview pre-screening gate} prior to full download: for each candidate video, we evaluate an unordered set of $N_{\text{pre}}$ sparse preview images ($5 \le N_{\text{pre}} \le 256$) independently through the shared single-frame heterogeneous hand-and-face perception policy (Section~3.3), requiring valid-wrist non-face-associated hands within active scale bounds while intentionally relaxing full-finger completeness so tool-occluded manipulation is not prematurely rejected. Letting $I_{\text{sup}}(r) \in \{0,1\}$ indicate a non-third-person operation-supporting frame and $I_{\text{1st}}(r), I_{\text{3rd}}(r) \in \{0,1\}$ denote geometric viewpoint votes, a video qualifies for full-stream ingestion only if:
\begin{equation}
    \frac{1}{N_{\text{pre}}} \sum_{r=1}^{N_{\text{pre}}} I_{\text{sup}}(r) \ge \theta_{\text{pre}} \quad\text{and}\quad \sum_{r=1}^{N_{\text{pre}}} I_{\text{1st}}(r) > \sum_{r=1}^{N_{\text{pre}}} I_{\text{3rd}}(r),
    \label{eq:preselect}
\end{equation}
with $\theta_{\text{pre}} = 0.16$. This gate discards roughly 96\% of deduplicated candidates upfront, admitting only $\sim 4\%$ into full-video acquisition and cryptographic content deduplication to form the 6.66M-video source corpus $\mathcal{V}$. Because an admitted long video $V \in \mathcal{V}$ still interleaves hands-on execution with narration and editing cuts, subsequent temporal curation extracts mutually non-overlapping sub-video intervals $\{(t_m^{\text{start}}, t_m^{\text{end}}, \mathcal{H}_m, \mathcal{Y}_m)\}_{m=1}^{M}$ satisfying $t_m^{\text{end}} - t_m^{\text{start}} > T_{\min}$ ($T_{\min} = 10.0\text{ s}$), paired with $1\text{ Hz}$ fused 2D hand trajectories $\mathcal{H}_m$ and structured semantic-visual annotations $\mathcal{Y}_m$.

Admitted intervals must satisfy three criteria: (1) resolved hand anatomy and adequate spatial scale within the active picture; (2) temporal and physical kinematic continuity, excluding fleeting appearances, conversational gestures, and stationary object holding; and (3) a unified first-person viewpoint. Accounting for camera-mount diversity in real-world web recordings, we unify head- or chest-mounted views (\texttt{first\_person}) and close-range unobstructed mounts positioned near the operator's torso along the line of sight (\texttt{overhead}) under a single first-person standard, while strictly excluding side, opposite, or distant third-person spectator views.

\subsection{Decode-Level Media Integrity and Neural Shot Partitioning}
Container metadata in web streams often masks letterbox/pillarbox padding, frame-duplication upsampling, or truncated tails. Prior to neural inference, each video undergoes decode-level verification: (1) full-resolution luma-edge cropping excludes black bars and requires an effective short side $\ge 720\text{ px}$ during screening (admitted clips are subsequently re-downloaded and recut at container short side $\ge 1080\text{p}$); (2) presentation timestamps (PTS) must be monotonic across the full duration with measured frame rate $\ge 24\text{ fps}$ and interval coefficient of variation $\text{CV} \le 0.5$; and (3) frame-duplication padding ratio must satisfy $\le 0.30$ over moving transitions, whereas static backgrounds on fixed mounts are deferred to hand-kinematics analysis.

Verified streams are partitioned into cut-free scenes via TransNet V2~\cite{soucek2020transnetv2}. Shot boundaries act as hard outer envelopes so camera cuts never span a clip. Fragmentary shots with duration $\le T_{\min}=10.0\text{ s}$ are pruned immediately, and only eligible long shots are sampled at $f_s = 1\text{ Hz}$ for downstream hand analysis.

\subsection{Heterogeneous Hand Fusion and Physical Kinematic Trimming}
Across $1\text{ Hz}$ frames within eligible shots, we run two complementary 2D hand models alongside a lightweight BlazeFace detector~\cite{bazarevsky2019blazeface}: (1) cascaded palm detection and cropped 21-landmark regression via MediaPipe Hands~\cite{zhang2020mediapipehands} (branch $\text{c}$), which offers fine joint precision in unobstructed views but drops entire hands when tools occlude the palm or rapid motion blurs the crop; and (2) the single-stage multi-scale full-frame hand detection and 2D 21-keypoint branch of WiLoR~\cite{potamias2025wilor} (branch $\text{d}$, running strictly 2D localization without 3D mesh reconstruction), which leverages global context for robust recall under occlusion and border truncation.

After per-joint confidence and border-margin masking within the active picture (masking out-of-bounds or low-confidence joints individually rather than discarding whole hands), we perform three-step same-frame fusion and geometric screening:
\begin{enumerate}
    \item \textbf{Cross-model merged-hand suppression and 1-to-1 association:} When the two branches report unequal hand counts, any single detection on the fewer-count side overlapping two distinct counterpart hands while splitting keypoint votes across both is vetoed as a merged-hand false positive. Remaining cascaded instances $i$ (landmark box $B_i^{\text{c}}$) and full-frame detections $j$ (detector box $B_j^{\text{d}}$) are scored by:
    \begin{equation}
        A(i, j) = \max\!\left(\operatorname{IoU}\!\left(B_i^{\text{c}}, B_j^{\text{d}}\right),\; 1 - \frac{\bar{d}_{ij}}{d_0}\right),
        \label{eq:affinity}
    \end{equation}
    where $\bar{d}_{ij}$ is the mean Euclidean distance in $[0,1]^2$ across $\ge 15$ shared valid joints ($d_0 = 0.20$). Gated pairs are matched greedily 1-to-1 in descending order of $A(i,j)$, while unmatched high-confidence detections are retained unilaterally.
    \item \textbf{Confidence synergy and 2D keypoint fusion:} For each joint $k \in \{0,\dots,20\}$ of a paired hand with valid 2D observations $(\mathbf{p}_k^{\text{c}}, c_k^{\text{c}})$ and $(\mathbf{p}_k^{\text{d}}, c_k^{\text{d}})$ (setting $c_k=0$ if missing on either side), fused 2D coordinates are computed via calibrated confidence weighting:
    \begin{equation}
        \hat{\mathbf{p}}_k = \frac{\alpha_{\text{c}} c_k^{\text{c}} \mathbf{p}_k^{\text{c}} + \alpha_{\text{d}} c_k^{\text{d}} \mathbf{p}_k^{\text{d}}}{\alpha_{\text{c}} c_k^{\text{c}} + \alpha_{\text{d}} c_k^{\text{d}}},
        \label{eq:fusion}
    \end{equation}
    with $\alpha_{\text{c}} = 1.20, \alpha_{\text{d}} = 1.0$. Consensus hands qualify under gentler thresholds to preserve rapid motion, while Eq.~\eqref{eq:fusion} seamlessly inherits unilateral high-confidence joints whenever the counterpart branch misses a frame or occluded finger.
    \item \textbf{Face-associated hand detachment and wrist-palm screening:} Hands geometrically belonging to detected in-frame faces are detached as presenter/bystander hands (passerby or small picture-in-picture faces alone do not reject genuine manipulation); remaining hands are screened by active-area ratio, wrist-below-palm entry orientation, and interval-level wrist-orientation consistency.
\end{enumerate}

Over $1\text{ Hz}$ fused observations, we apply \textbf{dual-track occlusion-tolerant continuity partitioning and physical kinematic trimming}. Independent short gap tolerances $\tau_{\text{wrist}} = 2\text{ s}$ (qualifying wrist absence) and $\tau_{\text{finger}} = 5\text{ s}$ (incomplete finger structure) bridge brief border exits and tool self-occlusions without fragmenting ongoing actions, while anchoring both interval endpoints on verified wrist-operation frames.

To physically excise passive stationary holding and boundary pauses, non-face-associated hands are projected into source pixel coordinates $\mathbf{x}_{t,k} \in [0, W_{\text{img}}]\times[0, H_{\text{img}}]$. Requiring $\ge 15$ valid joints, $\ge 3$ palm/MCP anchors ($\{0,5,9,13,17\}$), and distal visibility across all five digits, we define the instantaneous palm reference scale:
\begin{equation}
    s_t = \max\!\left(\|\mathbf{x}_{t,0} - \mathbf{x}_{t,9}\|_2,\; \|\mathbf{x}_{t,5} - \mathbf{x}_{t,17}\|_2\right).
    \label{eq:palm-scale}
\end{equation}
Establishing mutually unambiguous cross-frame matches strictly along selected operation-hand tracks (isolating unrelated bystander hands) over shared valid joints $\mathcal{K}$, a frame receives kinematic support from either adjacent-step displacement or sliding-window ($W = 5\text{ s}$) trajectory envelope diameter:
\begin{equation}
\begin{aligned}
    \max_{k \in \mathcal{K}} \|\mathbf{x}_{t,k} - \mathbf{x}_{t-1,k}\|_2 &> \max\!\left(\epsilon_{\text{px}},\, \eta_{\text{step}} s_{t-1}\right), \\
    \text{or}\quad \max_{u,v \in [t_0, t_1]} \max_{k \in \mathcal{K}} \|\mathbf{x}_{u,k} - \mathbf{x}_{v,k}\|_2 &> \max\!\left(2\epsilon_{\text{px}},\, \eta_{\text{drift}} s_{t_0}\right),
\end{aligned}
\label{eq:kinematics}
\end{equation}
where $|t_1 - t_0| < W = 5\text{ s}$, $\eta_{\text{step}} = 0.20$, $\eta_{\text{drift}} = 0.50$, and $\epsilon_{\text{px}} = 2.0\text{ px}$. Stationary runs ($\ge 3\text{ s}$) and unsupported low-motion runs ($\ge 6\text{ s}$) are excised along the timeline, and surviving sub-intervals must re-satisfy $T_{\min} = 10.0\text{ s}$.

\subsection{Multimodal Semantic Verification and Open-Vocabulary Annotation}
While front-end physical-geometric stages guarantee clear hand structure and active physical motion, unconstrained web footage still contains semantic non-target cases indistinguishable by 2D trajectories alone---specifically video games and 3D CG animations, driver-perspective vehicle steering, face-out-of-frame side/opposite third-person angles, split-screen composites, and empty-handed gesturing.

In the back-end stage, each candidate interval is represented as a temporally ordered multi-frame visual sequence sampled uniformly across its trimmed span and evaluated by OpenAI GPT-6 Astra (\texttt{gpt-6-astra}) under a unified structured verification protocol, with exact model identifiers, prompt digests, and token-usage telemetry recorded alongside each output. Clips are rejected if they depict: (1) virtual, animated, or video-game scenes (real hands drawing on paper or digital tablets are retained); (2) driver-perspective vehicle driving (mechanical/automotive repair is retained); (3) residual third-person spectator angles; or (4) multi-view split-screen/picture-in-picture layouts, pure empty-handed \texttt{gesture}, or visually indeterminate actions.

Verified clips retain their physically anchored microsecond boundaries and receive open-vocabulary English annotations alongside explicit visual-overlay attributes (Table~\ref{tab:annotation-schema}). For example, on a 46.0-second electronics repair segment, the annotation specifies: \texttt{caption}: \emph{``A person touches multimeter probes to a battery pack's connector contacts, adjusts the meter dial, then tests additional contacts.''}; \texttt{action}: \texttt{Test electrical contacts}; \texttt{object}: \texttt{Battery pack connector contacts}; \texttt{tool}: \texttt{Multimeter}; with all three overlay flags set to \texttt{false}.

\begin{table}[htbp]
\centering
\caption{Structured English semantic annotation and visual-overlay schema in \DatasetName{}}
\label{tab:annotation-schema}
\vspace{0.3em}
{\small
\begin{tabular}{lp{0.72\linewidth}}
\toprule
\textbf{Attribute} & \textbf{Definition and Generation Constraints} \\
\midrule
\texttt{caption} & \raggedright Concise single-sentence chronological English description (mean 16.1 words) summarizing visible hand-object interactions and stage transitions; state-change verbs require multi-frame visual evidence rather than static grasp guesses. \tabularnewline
\texttt{action} & \raggedright Single core English verb or verb phrase aligned with the primary manipulation in \texttt{caption}; \texttt{null} if indeterminate. \tabularnewline
\texttt{object} & \raggedright Single primary manipulated target distinct from tools (e.g., \texttt{potato} when cutting a potato with a knife); \texttt{null} if indeterminate. \tabularnewline
\texttt{tool} & \raggedright Single primary external tool used to execute the action (e.g., \texttt{screwdriver}, \texttt{multimeter}); \texttt{null} for bare-hand manipulation or unconfirmed cases (human hands and background items are never labeled as tools). \tabularnewline
Overlay attributes & \raggedright Explicit boolean flags for corner watermarks (\texttt{has\_watermark}), hardcoded subtitles (\texttt{has\_hardcoded\_subtitles}), and multi-view composites (\texttt{has\_multiple\_views}), enabling stratified subset selection. \tabularnewline
\bottomrule
\end{tabular}
}
\end{table}

To ensure scientific reproducibility and controlled ablation analysis at million-hour scale, \textbf{$1\text{ Hz}$ visual perception observations are represented decoupled from interval-retention and semantic criteria}. Each clip identifier binds cryptographic source digests, parent microsecond timestamps, $1\text{ Hz}$ fused hand trajectories, and prompt-hashed semantic outputs, allowing alternative geometric priors, occlusion tolerances, or kinematic thresholds (Section~5) to be evaluated identically over fixed underlying observations without confounding decoding or inference variance.

\section{The \DatasetName{} Dataset}
\subsection{Curation Policy and Comparison with Existing Datasets}
To serve both diversity-hungry VLA policy pretraining and overlay-sensitive video world models, we construct \DatasetName{} under two design principles: (1) \emph{hard physical-semantic admission}, strictly retaining unified first-person views (\texttt{first\_person} and close-range \texttt{overhead} while excluding all third-person clips), removing multi-view split-screen composites and pure empty-handed \texttt{gesture} segments, and enforcing short-side resolution $\ge 1080\text{p}$; and (2) \emph{explicit visual-overlay stratification}, providing \texttt{has\_watermark} and \texttt{has\_hardcoded\_subtitles} as boolean attributes while isolating the zero-watermark, zero-subtitle, single-view subset \DatasetName{}-Clean.

\begin{table}[htbp]
\centering
\caption{Comparison of \DatasetName{} with representative egocentric manipulation and web video datasets}
\label{tab:dataset-comparison}
\vspace{0.3em}
{\scriptsize
\setlength{\tabcolsep}{2.8pt}
\begin{tabular}{llrrlll}
\toprule
\textbf{Dataset} & \textbf{Source \& Setting} & \textbf{Hours} & \textbf{Clips} & \textbf{Res.} & \textbf{Hand Pose} & \textbf{Semantic Annotations} \\
\midrule
EPIC-KITCHENS~\cite{damen2022epickitchens} & Wearable (kitchen) & 100 & 90.0K & 1080p & None & Closed-set verb + noun \\
HOI4D~\cite{liu2022hoi4d} & Lab tabletop & 22 & 4.0K & 1080p & 3D mesh & 16 indoor object classes \\
Assembly101~\cite{sener2022assembly101} & Multi-view lab & 513$^{\dagger}$ & 101.0K & 720p+ & 3D pose & Toy assembly steps \\
SSv2~\cite{goyal2017something} & Crowdsourced clips & 245 & 220.8K & 240p+ & None & 174 template actions \\
Ego4D~\cite{grauman2021ego4d} & Multi-city wearable & 3,670 & ---$^{\ddagger}$ & 720p+ & Subset only & Narrations \& boxes (subset) \\
Ego-Exo4D~\cite{grauman2024egoexo4d} & Ego + exocentric & 221$^{\S}$ & 5.0K & 1080p+ & Subset 3D & Skilled commentary \& steps \\
EgoDex~\cite{hoque2025egodex} & HMD tabletop & 829 & 320.0K & 1080p & 3D skeletal & Prescribed tabletop tasks \\
HowTo100M~\cite{miech2019howto100m} & Uncurated web & 134,472$^{\P}$ & 136.6M & 240p+ & None & Unaligned ASR transcripts \\
\midrule
\textbf{\DatasetName{}-Clean (Ours)} & Open-web ego & \textbf{21,019} & \textbf{2.58M} & \textbf{$\ge$1080p} & \textbf{1 Hz 2D} & \textbf{Open-vocab + overlay-free} \\
\textbf{\DatasetName{} Core (Ours)} & Open-web ego & \textbf{39,400} & \textbf{5.12M} & \textbf{$\ge$1080p} & \textbf{1 Hz 2D} & \textbf{402K act / 404K obj / 21K tool} \\
\bottomrule
\end{tabular}
}
\par\vspace{0.25em}{\raggedright\scriptsize Notes: $^{\dagger}$Includes synchronized exocentric and egocentric streams; $^{\ddagger}$long-form recordings partitioned differently per benchmark; $^{\S}$egocentric streams only (1,286 h including exocentric views); $^{\P}$unfiltered viewpoints containing extensive narration and non-manipulation shots.\par}
\end{table}

Table~\ref{tab:dataset-comparison} contextualizes \DatasetName{} against prior benchmarks. Relative to dedicated wearable collections (EPIC-KITCHENS-100, Ego4D, EgoDex), \DatasetName{} maintains $\ge 1080\text{p}$ resolution, verified manipulation continuity, and $1\text{ Hz}$ hand keypoints while expanding total duration to 39,400.16 hours ($10.7\times$ Ego4D and $47.5\times$ EgoDex) across 686,914 real-world environments. Relative to uncurated instructional corpora such as HowTo100M, \DatasetName{} eliminates non-egocentric talking-head segments, editing cuts, and noisy ASR alignment via physical kinematic trimming and multimodal visual verification.

\subsection{Dataset Scale and Stratified Subsets}
Throughout all statistics, we distinguish deduplicated source videos from non-overlapping manipulation clips and accumulate durations strictly over disjoint intervals. As shown in Table~\ref{tab:zego-subsets}, the \DatasetName{} Core Release contains 5,117,314 high-definition manipulation clips totaling 39,400.16 hours from 686,914 unique source videos (7.45 clips per video, mean length 27.72 s). Within the overlay-annotated unified first-person subset (3,291,270 clips, 26,150.79 hours), 95.11\% are free of hardcoded subtitles and 84.09\% are free of watermarks; excluding split-screen views and gestures yields 3,093,438 single-view clips (24,648.06 hours), of which 2,583,138 clips (83.50\%, 21,018.85 hours) constitute the completely overlay-free subset \DatasetName{}-Clean.

\begin{table}[htbp]
\centering
\caption{Corpus scale and stratified subsets of \DatasetName{}}
\label{tab:zego-subsets}
\vspace{0.3em}
{\small
\setlength{\tabcolsep}{4pt}
\begin{tabular}{lrrrr}
\toprule
\textbf{Dataset Tier / Subset} & \textbf{Clips} & \textbf{Source Videos} & \textbf{Hours} & \textbf{Mean (s)} \\
\midrule
Multimodal-evaluated pool (incl.\ 3rd-person) & 5,698,502 & 820,454 & 43,620.04 & 27.56 \\
Unified 1st-person pool (3rd-person removed) & 5,323,702 & 717,236 & 40,957.60 & 27.70 \\
\textbf{\DatasetName{} Core Release (single-view)} & \textbf{5,117,314} & \textbf{686,914} & \textbf{39,400.16} & \textbf{27.72} \\
\quad Overlay-annotated unified 1st-person set & 3,291,270 & 621,408 & 26,150.79 & 28.60 \\
\quad Overlay-annotated single-view set & 3,093,438 & 595,596 & 24,648.06 & 28.68 \\
\quad \textbf{Overlay-free subset (\DatasetName{}-Clean)} & \textbf{2,583,138} & \textbf{430,867} & \textbf{21,018.85} & \textbf{29.29} \\
\bottomrule
\end{tabular}
}
\end{table}

\subsection{Video Signal Characteristics and Temporal Distributions}
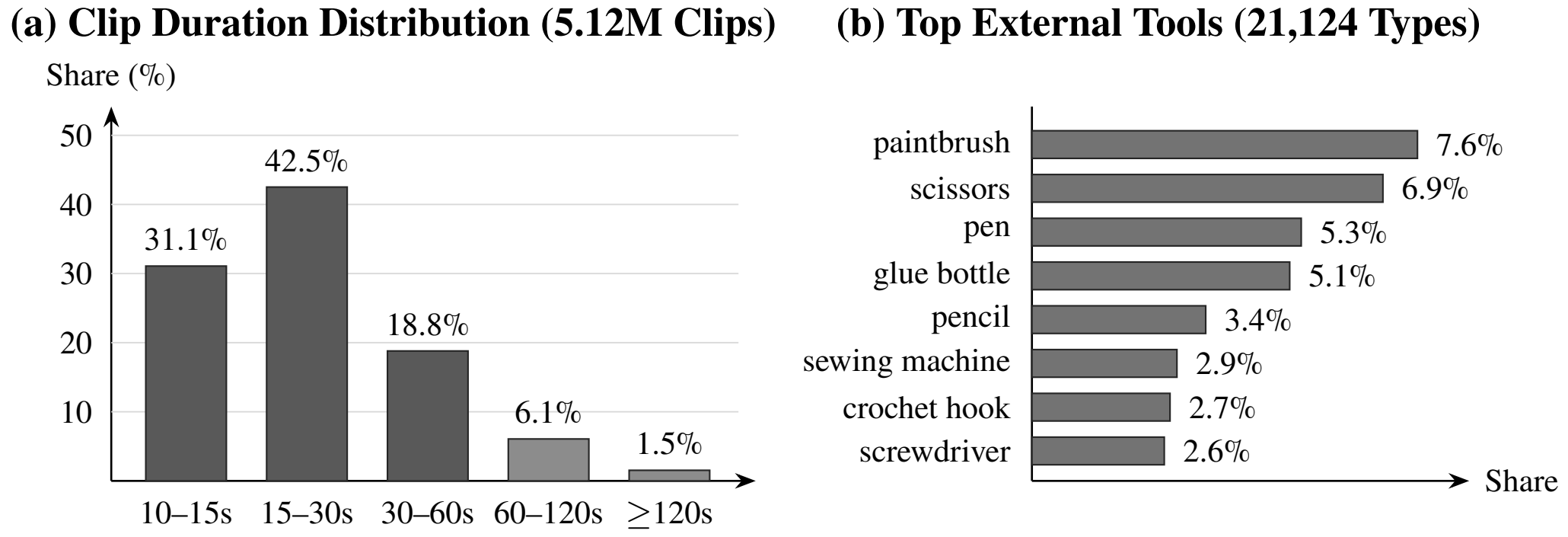
\begin{figure}[htbp]
\centering
\begin{tikzpicture}[font=\footnotesize]
\begin{scope}[xshift=0cm]
    \node[font=\small\bfseries] at (3.0, 3.95) {(a) Clip Duration Distribution (5.12M Clips)};
    \draw[-{Stealth[length=1.8mm]}, line width=0.5pt] (0.55,0) -- (6.15,0);
    \draw[-{Stealth[length=1.8mm]}, line width=0.5pt] (0.55,0) -- (0.55,3.25) node[above, font=\scriptsize] {Share (\%)};
    \foreach \y/\lbl in {0.6/10, 1.2/20, 1.8/30, 2.4/40, 3.0/50} {
        \draw[black!15, line width=0.35pt] (0.55,\y) -- (6.0,\y);
        \node[left, font=\scriptsize] at (0.52,\y) {\lbl};
    }
    \filldraw[fill=black!65, draw=black!85] (0.85,0) rectangle (1.55,1.865);
    \node[above, font=\scriptsize] at (1.20,1.865) {31.1\%};
    \node[below, font=\scriptsize] at (1.20,-0.04) {10--15s};

    \filldraw[fill=black!65, draw=black!85] (1.90,0) rectangle (2.60,2.551);
    \node[above, font=\scriptsize] at (2.25,2.551) {42.5\%};
    \node[below, font=\scriptsize] at (2.25,-0.04) {15--30s};

    \filldraw[fill=black!65, draw=black!85] (2.95,0) rectangle (3.65,1.128);
    \node[above, font=\scriptsize] at (3.30,1.128) {18.8\%};
    \node[below, font=\scriptsize] at (3.30,-0.04) {30--60s};

    \filldraw[fill=black!45, draw=black!85] (4.00,0) rectangle (4.70,0.364);
    \node[above, font=\scriptsize] at (4.35,0.364) {6.1\%};
    \node[below, font=\scriptsize] at (4.35,-0.04) {60--120s};

    \filldraw[fill=black!45, draw=black!85] (5.05,0) rectangle (5.75,0.092);
    \node[above, font=\scriptsize] at (5.40,0.092) {1.5\%};
    \node[below, font=\scriptsize] at (5.40,-0.04) {$\ge$120s};
\end{scope}

\begin{scope}[xshift=6.80cm]
    \node[font=\small\bfseries] at (2.85, 3.95) {(b) Top External Tools (21,124 Types)};
    \draw[-{Stealth[length=1.8mm]}, line width=0.5pt] (1.75,0) -- (5.55,0) node[right, font=\scriptsize] {Share};
    \draw[line width=0.5pt] (1.75,0) -- (1.75,3.25);
    \foreach \idx/\name/\val/\w in {
        7/paintbrush/7.6\%/3.35,
        6/scissors/6.9\%/3.05,
        5/pen/5.3\%/2.34,
        4/glue bottle/5.1\%/2.24,
        3/pencil/3.4\%/1.51,
        2/sewing machine/2.9\%/1.26,
        1/crochet hook/2.7\%/1.20,
        0/screwdriver/2.6\%/1.15
    } {
        \pgfmathsetmacro{\yb}{0.14 + \idx * 0.38}
        \pgfmathsetmacro{\yt}{\yb + 0.24}
        \pgfmathsetmacro{\ym}{\yb + 0.12}
        \filldraw[fill=black!55, draw=black!85] (1.75,\yb) rectangle ({1.75+\w},\yt);
        \node[left, font=\scriptsize] at (1.70,\ym) {\name};
        \node[right, font=\scriptsize] at ({1.78+\w},\ym) {\val};
    }
\end{scope}
\end{tikzpicture}
\caption{Empirical distributions of \DatasetName{} Core: (a) clip duration distribution (median 19.0 s, mean 27.7 s, 7.60\% exceeding 60 s); (b) leading tools within the tool-assisted subset (38.77\% of clips across 21,124 tool types), where the top eight tools account for only 36.5\%, reflecting pronounced open-world long-tail diversity.}
\label{fig:zego-distributions}
\end{figure}

Table~\ref{tab:zego-media} and Figure~\ref{fig:zego-distributions}(a) summarize visual signal specifications across all 5.12M clips alongside frame-level decode statistics over $N=157$ stratified random clips. Every clip satisfies short-side resolution $\ge 1080\text{p}$ (85.99\% $1920\times 1080$, 9.25\% $3840\times 2160$, 1.83\% $2560\times 1440$, and 2.93\% ultra-wide or micro-cropped $\ge 1080\text{p}$ formats), with mean bitrate 7.51 Mbps (median 5.94 Mbps). Sampled clips are 100.00\% standard \texttt{H.264 (yuv420p)} SDR with intact frame decoding, minimum frame rate 30.0 fps, and timestamp interval $\text{CV} \le 0.5$; after luma black-border exclusion (present in 6.37\% of clips), 94.90\% retain effective short side $\ge 1080\text{ px}$ and 100.00\% satisfy $\ge 720\text{ px}$, while 98.09\% exhibit native non-duplicated motion (duplication ratio $\le 0.30$). Clip durations span 10.01 s to multi-minute episodes (median 19.00 s, mean 27.72 s, $P_{90}=52.00\text{ s}$), including 388,959 long-horizon clips (7.60\%) exceeding 60 s.

\begin{table}[htbp]
\centering
\caption{Visual resolution, bitrate, and temporal signal statistics of \DatasetName{}}
\label{tab:zego-media}
\vspace{0.3em}
{\footnotesize
\setlength{\tabcolsep}{3.5pt}
\begin{tabular}{lr}
\toprule
\textbf{Visual Signal \& Temporal Property} & \textbf{Distribution / Proportion} \\
\midrule
Video resolution ($N=5,117,314$) & 1080p: 85.99\%, 4K: 9.25\%, 2K: 1.83\%, other $\ge$1080p: 2.93\% \\
Duration quantiles (seconds) & min 10.01, $P_{25}$=14.0, $P_{50}$=19.0, mean 27.72, $P_{90}$=52.0 \\
Video bitrate quantiles (Mbps) & $P_{10}$=3.51, $P_{50}$=5.94, mean 7.51, $P_{90}$=12.85 \\
Color encoding \& frame integrity (sampled $N=157$) & \texttt{H.264 (yuv420p)} SDR: 100.00\% (corrupt-free) \\
Frame rate \& PTS interval stability & $\ge 30.0\text{ fps}$ and $\text{CV}\le 0.5$: 100.00\% \\
Effective short side after border exclusion & $\ge 720\text{ px}$: 100.00\%; $\ge 1080\text{ px}$: 94.90\% \\
Native non-duplicated motion (duplication $\le 0.30$) & 98.09\% \\
\bottomrule
\end{tabular}
}
\end{table}

\subsection{Long-Tail Semantic Diversity, Release Protocol, and Ethics}
Across 5.12 million clips, English captions average 16.1 words and span 401,576 unique action phrases, 404,440 manipulated object expressions (99.92\% non-null coverage), and 21,124 distinct external tools. Bare-hand dexterous interactions account for 61.23\% (3.13M clips, e.g., folding, kneading, assembling, sorting, unpacking), while tool-assisted manipulations account for 38.77\% (1.98M clips; Figure~\ref{fig:zego-distributions}(b)), with over 63\% of tool-assisted episodes distributed across long-tail instruments such as tweezers, hot glue guns, utility knives, soldering irons, wrenches, and knitting needles.

\paragraph{Release Format, Licensing, and Privacy.}
Following established academic practices for web-sourced video benchmarks (e.g., Kinetics, HowTo100M~\cite{miech2019howto100m}, and Panda-70M~\cite{chen2024panda70m}), \DatasetName{} is released for non-commercial academic research as structured metadata and derived geometric-semantic packages indexed by \texttt{<clip\_id>}: (1) origin platform identifiers, exact microsecond start/end timestamps, resolution digests, and standardized recut reproduction scripts (\texttt{.meta.json}); (2) $1\text{ Hz}$ fused 2D 21-joint coordinates and confidence weights (\texttt{.hands.json}); and (3) version- and prompt-hash-bound semantic annotations and visual-cleanliness flags (\texttt{.annotation.json}). Raw third-party video files are not publicly re-hosted or sold commercially; moreover, face-associated hand detachment and third-person rejection remove camera-facing talking-head segments by construction, and a permanent creator opt-out and takedown channel is maintained to respect creator rights and individual privacy.

\section{Experiments and Ablation Analysis}
We evaluate our multi-stage curation methodology at two complementary scales: (1) controlled stage-wise convergence and module-level ablations on 200 randomly sampled, unselected open-web videos totaling 117.55 hours (423,191 seconds; 44,071 shot segments), and (2) large-scale cascade analysis across 2.07 million hours of preview-admitted web videos alongside candidate precision evaluation over a uniformly hash-sampled set of 366,554 front-end candidate clips.

\subsection{Stage-Wise Convergence and Filtering Efficiency}
Table~\ref{tab:stage-funnel} tracks sample convergence across the 117.55-hour unselected evaluation corpus. Raw instructional and everyday recordings exhibit severe editing fragmentation (44,071 shots, mean length $<10\text{ s}$). Enforcing $T_{\min}=10.0\text{ s}$ immediately after neural shot partitioning eliminates 86.03\% of rapid cuts (retaining 6,155 eligible shots) and reduces $1\text{ Hz}$ hand-inference frames by 28.57\% (from 423,194 to 302,289 frames).

\begin{table}[htbp]
\centering
\caption{Stage-wise curation convergence on 200 unselected web videos (117.55 hours)}
\label{tab:stage-funnel}
\vspace{0.3em}
{\small
\setlength{\tabcolsep}{6pt}
\begin{tabular}{lrrr}
\toprule
\textbf{Front-End Curation Stage} & \textbf{Retained Intervals / Clips} & \textbf{Hours} & \textbf{Share of Raw} \\
\midrule
Raw input source videos (200 videos) & 44,071 (all shots) & 117.55 & 100.00\% \\
Neural shot partitioning \& $>10\text{ s}$ filter & 6,155 (eligible shots) & 83.97 & 71.44\% \\
Heterogeneous hand fusion \& continuity split & 1,443 (hand intervals) & 12.56 & 10.69\% \\
Physical kinematic trimming (final candidates) & 1,472 (dynamic clips) & 10.37 & 8.82\% \\
\bottomrule
\end{tabular}
}
\end{table}

Within eligible shots, heterogeneous hand fusion, face-hand detachment, and dual-track continuity narrow retention to 12.56 hours; physical kinematic trimming removes passive holding and boundary pauses, yielding 1,472 dynamic manipulation candidates across 83 videos (10.37 hours, mean 25.37 s, 8.82\% of raw duration). Overall, front-end curation eliminates 91.18\% of non-target duration and reduces downstream vision-language model invocations by 96.66\% relative to scoring all shots and 94.78\% relative to fixed 15-second windows (28,213 windows).

\subsection{Ablation on Heterogeneous Hand Perception Fusion}
Holding shot boundaries, face-hand geometry, dual-track tolerances, and kinematic trimming fixed on the 117.55-hour corpus, Table~\ref{tab:fusion-ablation} compares cascaded palm-landmark regression alone~\cite{zhang2020mediapipehands}, single-stage full-frame detection and keypoint estimation alone~\cite{potamias2025wilor}, and our instance association and confidence-weighted fusion.

\begin{table}[htbp]
\centering
\caption{Ablation comparing individual hand perception models against complementary fusion (117.55-hour corpus)}
\label{tab:fusion-ablation}
\vspace{0.3em}
{\small
\setlength{\tabcolsep}{4.5pt}
\begin{tabular}{lrrrr}
\toprule
\textbf{Hand Perception Configuration} & \textbf{Videos} & \textbf{Valid Frames} & \textbf{Clips} & \textbf{Retained Hours} \\
\midrule
Cascaded palm-landmark regression only~\cite{zhang2020mediapipehands} & 44 & 9,283 & 521 & 3.07 (2.61\%) \\
Single-stage full-frame detection-keypoint only~\cite{potamias2025wilor} & 76 & 25,302 & 1,194 & 7.94 (6.75\%) \\
\textbf{Same-frame association \& fusion (Ours)} & \textbf{83} & \textbf{33,813} & \textbf{1,472} & \textbf{10.37 (8.82\%)} \\
\bottomrule
\end{tabular}
}
\end{table}

Individual models exhibit pronounced temporal fragmentation under real-world occlusion and motion blur, retaining only 3.07 hours (cascaded) and 7.94 hours (full-frame). Complementary fusion expands continuous manipulation yield to 10.37 hours (1,472 clips)---improvements of 238.5\% and 30.7\%, respectively. Because consecutive misses exceeding $\tau_{\text{wrist}}=2\text{ s}$ sever ongoing episodes into sub-10-second fragments that are discarded entirely, bridging brief single-model dropouts through counterpart observations recovers complete manipulation trajectories rather than merely concatenating disjoint detections.

\subsection{Ablation on Dual-Track Occlusion Tolerance and Kinematic Trimming}
Table~\ref{tab:temporal-ablation} isolates dual-track occlusion tolerance ($\tau_{\text{wrist}}=2\text{ s}, \tau_{\text{finger}}=5\text{ s}$) and palm-scale-normalized kinematic trimming. Enforcing strict zero-gap continuity ($\tau_{\text{wrist}}=\tau_{\text{finger}}=0$) reduces retained duration by 44.1\% (10.37 h to 5.79 h; mean length 25.37 s to 20.23 s), confirming that brief partial occlusions are intrinsic to natural hand-object contact. Conversely, disabling kinematic trimming inflates candidate volume to 12.56 hours across 87 videos by retaining 2.19 hours (17.43\%) of passive object display and boundary pauses; enforcing Eq.~\eqref{eq:kinematics} eliminates 4 purely static videos and refines mean clip length from 31.34 s to 25.37 s.

\begin{table}[htbp]
\centering
\caption{Ablation on dual-track occlusion tolerance and physical kinematic trimming (117.55-hour corpus)}
\label{tab:temporal-ablation}
\vspace{0.3em}
{\small
\setlength{\tabcolsep}{4.5pt}
\begin{tabular}{lrrrr}
\toprule
\textbf{Temporal \& Kinematic Configuration} & \textbf{Videos} & \textbf{Clips} & \textbf{Hours} & \textbf{Mean Length (s)} \\
\midrule
Strict zero-gap ($\tau_{\text{wrist}}=0,\tau_{\text{finger}}=0$) & 66 & 1,031 & 5.79 & 20.23 \\
Without physical kinematic trimming & 87 & 1,443 & 12.56 & 31.34 \\
\textbf{Dual-track tolerance + kinematic trim (Ours)} & \textbf{83} & \textbf{1,472} & \textbf{10.37} & \textbf{25.37} \\
\bottomrule
\end{tabular}
}
\end{table}

\subsection{End-to-End Cascade Scarcity and Candidate Precision Verification}
Evaluating the full production cascade reveals the essential duality between aggressive upstream rejection of low-likelihood videos and high-recall intra-video recovery (Section~5.2). At the entry gate, across $\sim 230\text{M}$ candidate videos queried via $\sim 900\text{K}$ multilingual search keywords, sparse preview pre-screening (Eq.~\eqref{eq:preselect}) discards roughly 96\% of deduplicated web candidates upfront, admitting only $\sim 4\%$ into full-stream ingestion---yielding 6,659,815 deduplicated source videos totaling 2,074,386.0 hours (mean length 18.69 minutes). Even within these preview-qualified videos, full-stream media integrity checks, shot partitioning, heterogeneous hand fusion, and kinematic trimming eliminate a further 1,759,857.4 hours (84.84\%) of narration, B-roll, and stationary intervals, isolating 314,528.6 hours (15.16\%) of non-overlapping dynamic manipulation candidates (rising from 8.82\% on the unscreened 200-video cohort in Section~5.1).

\begin{table}[htbp]
\centering
\caption{Multimodal semantic verification breakdown on uniformly hash-sampled front-end candidates ($N=366{,}554$)}
\label{tab:llm-precision}
\vspace{0.3em}
{\small
\setlength{\tabcolsep}{5pt}
\begin{tabular}{lrrrr}
\toprule
\textbf{Multimodal Verification Category} & \textbf{Clips} & \textbf{Clip Share} & \textbf{Hours} & \textbf{Duration Share} \\
\midrule
\textbf{Verified unified 1st-person manipulation} & \textbf{328,821} & \textbf{89.71\%} & \textbf{2,367.14} & \textbf{89.79\%} \\
Face-out-of-frame 3rd-person spectator views & 30,153 & 8.23\% & 215.58 & 8.18\% \\
Driver-perspective vehicle steering & 2,946 & 0.80\% & 23.13 & 0.88\% \\
Video games, animation, and 3D CG rendering & 2,087 & 0.57\% & 13.76 & 0.52\% \\
Ambiguous or insufficient visual evidence & 2,547 & 0.69\% & 16.80 & 0.64\% \\
\bottomrule
\end{tabular}
}
\end{table}

Table~\ref{tab:llm-precision} reports multimodal verification outcomes (GPT-6 Astra) across a uniformly hash-sampled subset of 366,554 front-end candidates (2,636.41 hours): without invoking any semantic model, front-end physical-geometric candidates achieve 89.71\% unified first-person manipulation precision (89.79\% by duration), while semantic gating removes the residual 10.29\% non-target tail---chiefly face-out-of-frame third-person views (8.23\%) alongside driving, virtual/CG rendering, and ambiguous clips (2.06\% combined). Multiplying survival rates across the entire cascade---$\sim 4\%$ sparse preview admission $\times 15.16\%$ in-stream temporal-kinematic retention $\times 89.79\%$ multimodal verification---yields a compound end-to-end yield of only $\approx 0.54\%$, dropping further to $\approx 0.44\%$ when enforcing single-view watermark- and subtitle-free cleanliness (\DatasetName{}-Clean; fewer than 1 part in 200 of raw discovered content). This stark compound scarcity demonstrates that while open-web repositories contain vast manipulation diversity in aggregate, genuinely clean, cut-free, high-definition egocentric manipulation is extraordinarily rare and deeply entangled with editing noise---requiring both selective upstream pre-screening to reject low-probability videos and occlusion-resilient heterogeneous fusion (Section~5.2) to preserve valid episodes within admitted streams.

\paragraph{Multimodal Token Scale and Clean-Hour Cost Efficiency.}
De-duplicating request-level usage logs across batched multi-clip API calls ($K \in [3, 10]$ clips per request; $\approx 9.62\text{B}$ cumulative input tokens and $\approx 10.1\text{B}$ total tokens across all curated batches), evaluating one front-end candidate under our full semantic and overlay verification protocol with GPT-6 Astra consumes an amortized mean of 1,808 input tokens (1,255 visual and clip-context tokens plus 553 amortized system-prompt tokens, of which 509 hit automatic prefix caching) and 85 output tokens---making token expenditure 95.5\% input-dominated (1,893 total tokens/clip). Benchmarked against our strictest single-view overlay-free subset \DatasetName{}-Clean and accounting for all candidates filtered out during multimodal verification and overlay screening (70.41\% net candidate-to-Clean clip yield), distilling 1 hour of final qualified clean video (122.90 clean clips, requiring $\approx 174.5$ evaluated front-end candidates) consumes $\approx 3.15\times 10^5$ (0.315M) input tokens (0.227M uncached plus 0.089M prefix-cached). Valued at official standard OpenAI GPT-6 Astra API input rates (\$10.00 / 1M uncached input tokens; \$1.00 / 1M prefix-cached input tokens), multimodal verification and open-vocabulary annotation cost $\approx \$2.36$ in input tokens per hour of final overlay-free manipulation video ($\approx \$3.15/\text{h}$ without prefix caching). Without our front-end physical-geometric cascade, scoring raw discovered web videos in fixed 15-second windows at $\approx 0.44\%$ clean target density would require nearly $10^8$ input tokens ($\sim \$1{,}000$) per clean hour, underscoring more than two orders of magnitude reduction in multimodal token expenditure.

\section{Limitations and Conclusion}
\subsection{Limitations and Future Work}
Despite significant advances in scale, environmental diversity, and kinematic continuity, several limitations merit future investigation. First, our curation framework and released \texttt{.hands.json} trajectories provide fused 2D 21-keypoint observations in image space rather than metric 3D hand-object meshes~\cite{romero2017mano}, depth maps, or calibrated camera extrinsics; lifting high-definition clips and 2D anchors to metric 3D hand-object reconstructions is a natural extension. Second, explicit watermark and hardcoded-subtitle boolean attributes currently cover 62.6\% of the 39,400.16-hour Core Release (24,648.06 hours of overlay-annotated single-view clips, from which the 21,018.85-hour \DatasetName{}-Clean subset is derived), while the remaining portion enforces unified first-person single-view manipulation admission without fine-grained overlay flags. Third, this work focuses on curation methodology, empirical corpus characterization, and open dataset construction; systematic downstream scaling experiments for world-action models and physical robot transfer are left to dedicated follow-up studies.

\subsection{Conclusion}
We have presented a multi-stage curation and automatic annotation methodology for mining egocentric human manipulation videos from unconstrained web sources, together with the \DatasetName{} dataset. Coupling sparse preview pre-screening, decode-level media verification, neural shot partitioning, same-frame instance association and confidence-weighted 2D keypoint fusion across heterogeneous hand models, dual-track occlusion tolerance with palm-scale-normalized kinematic trimming, and multimodal semantic verification discards $\sim 96\%$ of low-likelihood candidates upfront while resolving shot fragmentation, occlusion dropouts, and non-manipulative noise within admitted streams---distilling clean episodes across an end-to-end cascade yield of only $\approx 0.54\%$. \DatasetName{} provides 5,117,314 high-definition manipulation clips (39,400.16 hours) across 686,914 source videos and over 400,000 action and object expressions---including the 21,018.85-hour overlay-free subset \DatasetName{}-Clean---as open data infrastructure for embodied representation learning, VLA models, and physical world modeling.

\end{document}